\documentclass{article}

\usepackage{PRIMEarxiv}

\usepackage[utf8]{inputenc} 
\usepackage[T1]{fontenc}    
\usepackage{hyperref}       
\usepackage{url}            
\usepackage{booktabs}       
\usepackage{amsfonts}       
\usepackage{nicefrac}       
\usepackage{microtype}      
\usepackage{lipsum}
\usepackage{fancyhdr}       
\usepackage{graphicx}       
\usepackage{subcaption}     
\usepackage{makecell}       
\graphicspath{{figures/}{./}}

\usepackage{natbib}
\usepackage{amsmath}        
\usepackage{amssymb}
\usepackage{amsthm}         
\usepackage{xcolor}         
\usepackage{multirow}       
\usepackage{colortbl}       
\usepackage{algorithm}      
\usepackage{algpseudocode}  
\algrenewcommand\algorithmicrequire{\textbf{Input:}}
\algrenewcommand\algorithmicensure{\textbf{Output:}}
\definecolor{mylightblue}{rgb}{0.9, 0.9, 1.0}

\usepackage{tikz}
\usetikzlibrary{positioning, arrows.meta, fit, calc}

\title{Revealing Safety-Critical Scenarios for UTM via Transformer
}

\author{
  Huaze Tang\thanks{Contributed equally.} \thanks{Work done while interning at Meituan.} \\
  Tsinghua University \\
   \And
  Bill Zeng\footnotemark[1] \\
  Meituan \\
   \And
  Chao Wang\footnotemark[2] \\
  Tsinghua University \\
   \AND
  Zhenpeng Shi\footnotemark[2] \\
  Tsinghua University \\
   \And
  Qian Zhang \\
  Meituan \\
   \And
  Wenbo Ding\thanks{Corresponding Author. \url{ding.wenbo@sz.tsinghua.edu.cn}} \\
  Tsinghua University \\
}

\begin{document}
\maketitle

\begin{abstract}
Unmanned Traffic Management (UTM) systems are cloud-based platforms designed to manage and coordinate multiple aerial vehicles remotely. UTM systems are safety-critical which cannot tolerate failures like crash or collision. To reveal latent vulnerabilities, there are neither optimal failure-exposing demonstrations nor clear reward signals. Additionally, UTM's self-healing capability introduces the ``long-tail effect'' of critical failures. We propose framing UTM vulnerability discovery as a sequence modeling problem amenable to transformer-based RL architectures. Our approach leverages attention mechanisms to directly model the relationship among system states, and predict optimal actions. Our framework introduces a Policy Model that generates targeted test scenarios and an Action Sampler that enforces domain constraints. We use a risk-based reward function to guide exploration. Through extensive evaluation on a 700-hour simulation study, we demonstrate an 8$\times$ improvement in vulnerability discovery efficiency compared to expert-guided testing. It also discovers critical edge cases that traditional methods have missed.
\end{abstract}

\keywords{Unmanned Traffic Management \and Safety-Critical Testing \and Decision Transformer \and Offline Reinforcement Learning \and Scenario Generation}

\section{Introduction}
The rapid emergence of the low-altitude economy has necessitated robust Unmanned Traffic Management (UTM) systems \citep{FAA_2023}. UTM performs centralized traffic control among aerial vehicles. Most system failures in UTM are intolerable, like crashing, collisions, airspace violations\citep{kopardekar2014unmanned,kopardekar2016unmanned}, making it vital to discover potential vulnerability scenarios during UTM iteration and deployment. In this context, a scenario is defined as a temporal segment of operational trajectories and environmental information of intelligent agents managed by the UTM system \citep{TianMOSATfindingsafetyviolationsautonomousdrivingsystemsusingmultiobjectivegeneticalgorithm2022, zhongSurveyScenarioBasedTesting2021a}.

\begin{figure*}[t]
\begin{center}
\includegraphics[height=4cm]{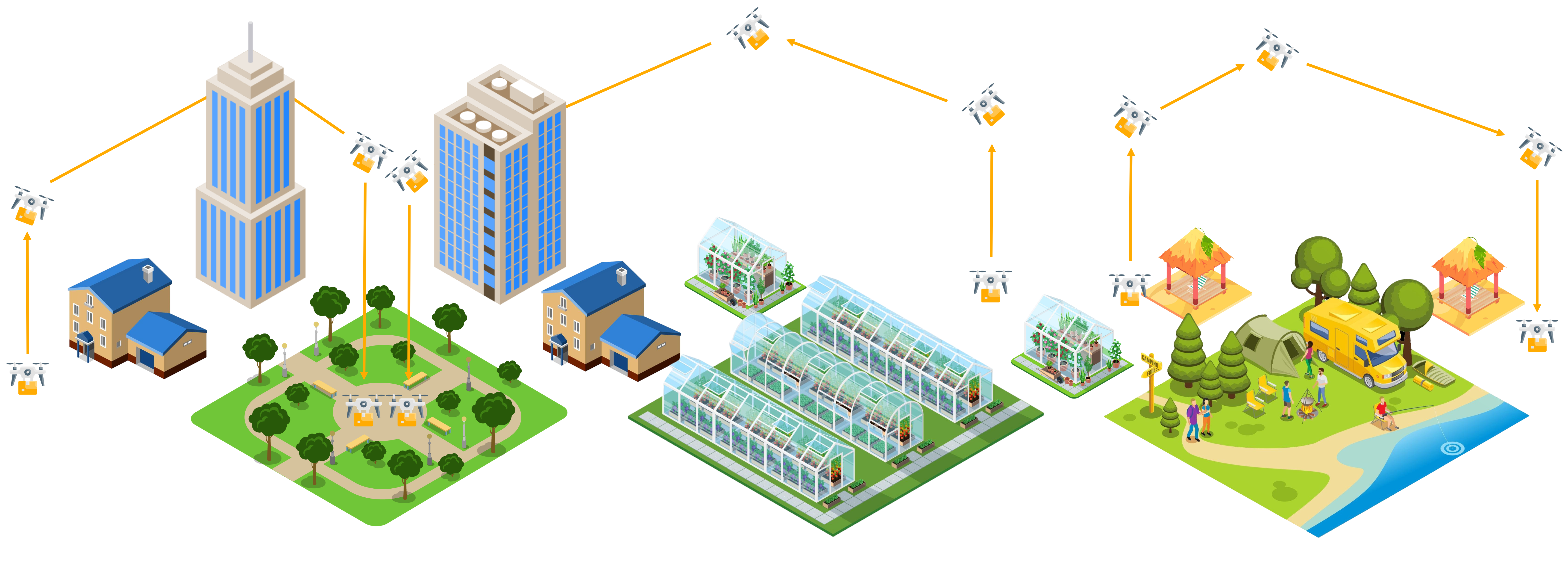}
\end{center}
\caption{\textbf{Overview of the operational environment in Unmanned aircraft system Traffic Management (UTM) of System-Under-Test (SUT).}  The UTM system operates in a variety of environments, including urban, suburban, and rural areas. Each setting poses distinct challenges, such as high-density air traffic in urban regions and limited infrastructure in rural areas, requiring strong management and coordination strategies. Rapid fault detection across these diverse scenarios is essential for maintaining safety and preventing catastrophic failures in real-world deployments.}
\label{fig:UAV_System_Overview}
\end{figure*}

UTMs are developed with self-healing functionality \citep{l.gladenceSwarmIntelligenceDisaster2021}. This feature helps prevent system failures automatically. However, it creates a challenge for testing. Most historical data contains only safe operational trajectories. The challenge of ``long-tail effect'' is thereby introduced. Moreover, these edge cases often emerge from subtle multi-agent runtime interaction \citep{wedadalawadUnmannedAerialVehicle2023} rather than simple component failures, suffering from a lack of optimal demonstrations or clear reward signals.

To address these challenges, we reformulate UTM vulnerability discovery as a sequence modeling problem amenable to modern transformer-based \citep{chen2021decisiontransformer} reinfocement learning (RL) architectures, as detailed in Section \ref{sec:transformer}. Our approach leverages the transformer's powerful attention mechanism \citep{VaswaniAttentionAllyouNeed2017} to directly model the relationship between system states, actions (anomaly perturbation), and outcomes in latent space \citep{leeMultigameDecisionTransformers2022}, as well as directly predict actions. This allows us to generate targeted test scenarios while avoiding both distributional barriers \citep{fuContextImitationLearning2024} and the shortage of expert demonstration data \citep{bhargavaWhenshouldwe2023}.

Our framework consists of two key components: a Policy Model (PM) that learns to generate targeted test scenarios based on both historical operational data and real-time system states, and an Action Sampler (AS) that enforces domain constraints to ensure physical plausibility. The PM captures complex temporal dependencies and inter-agent interactions through its attention mechanism, enabling it to identify patterns that may lead to system vulnerabilities.
We evaluate this approach in a large-scale simulation environment spanning 700 hours of testing across diverse operational contexts. Results demonstrate that our framework achieves over 8 times higher efficiency in vulnerability discovery compared to expert-guided exploitation, while also uncovering critical edge cases that traditional methods failed to detect. This significant improvement in testing efficiency and effectiveness showcases the potential of transformer-based approaches in fortifying mission-critical systems through intelligent scenario generation.

\section{Related Works}

\subsection{Mission Critical System Testing}
Testing autonomous systems, especially those operating in complex and dynamic environments like UTM systems and autonomous vehicles, requires comprehensive evaluation under diverse and challenging scenarios. While these domains may differ in their specific applications, they share common challenges in scenario generation: the need to efficiently explore vast parameter spaces, identify safety-critical cases, and maintain scenario plausibility.

In the field of autonomous driving scenario generation, researchers have proposed various approaches to generate challenging test scenarios. \citet{okellyScalableEndtoEndAutonomous2019} presented a scalable testing framework based on rare-event simulation based on GAIL and cross-entropy sampling. \citet{dingLearningCollideAdaptive2020} formulated scenario generation as a reinforcement learning problem, proposing an adaptive editing framework that constructs safety-critical scenarios through sequential editing operations (e.g., adding or modifying agents). In their subsequent work, \citet{dingRealGenRetrievalAugmented2024} further explored a retrieval-augmented generation (RAG) approach, which generates new scenarios by retrieving and combining features from existing scenarios, achieving better controllability. \citet{liuSafetyCriticalScenarioGeneration2024} proposed a reinforcement learning-based editing framework that supports more flexible scenario construction.

\subsection{Scenario Generation via Reinforcement Learning}
Learning to generate plausible trajectories from historical data is fundamental to both behavior prediction and scenario generation. Early works in inverse reinforcement learning (IRL) \citep{ngAlgorithmsInverseReinforcement2000} aim to recover the underlying reward function from expert demonstrations, with maximum entropy IRL \citep{ziebartMaximumEntropyInverse2008} providing a probabilistic framework that addresses the ambiguity in expert behaviors. To scale to high-dimensional problems, GAIL \citep{hoGenerativeAdversarialImitation2016} and its variants like AIRL \citep{fuLearningRobustRewards2018} employ adversarial training to directly learn policies through explicit reward recovery. Recently, transformer-based approaches have shown promising results by reformulating sequential decision making as a sequence modeling problem. Decision Transformer \citep{chen2021decisiontransformer} demonstrates that trajectory generation can be achieved through autoregressive prediction conditioned on desired returns, while Trajectory Transformer \citep{jannerOfflineReinforcementLearning2021a} treats both states and actions as tokens in a unified sequence model. These approaches provide different perspectives on trajectory modeling: IRL methods focus on understanding the underlying decision-making process \citep{aroraSurveyInverseReinforcement2020}, while transformer-based methods leverage the power of attention mechanisms to capture long-range dependencies in behavioral patterns.

Their complementary strengths suggest potential benefits in combining both frameworks for more effective trajectory and scenario generation. In this work, we hybrid the innovation of IRL that learn the world characteristics instead of expert behavior, along with the capability in latent space of Transformer architectures, in favor of extend the boarder of UTM testing scenario generation.

\section{Problem Analysis}

\subsection{UTM Characteristics}
UTM is a system to ensure safe and efficient operation of multiple UAVs (Unmanned Aerial Vehicles) in shared airspace \citep{kopardekar2014unmanned}, without requiring human air traffic controllers to manage each UAV directly. Detailed concepts are further illustrated in Appendix.\ref{Appendix.UTM}. The UTM system needs to process high-dimensional data inputs of multiple UAVs, as well as collision detection and route optimization simultaneously.

The objective of modern UTM vulnerability discovery demands active generation of failure-triggering sequences. However, learning must occur from a dataset dominated by safe trajectories, with only sporadic and potentially suboptimal examples of failures. This presents a unique imbalanced learning challenge.

Given $N$ UAVs, the UTM system evolves according to its internal protocols, generating operational trajectories $\boldsymbol{\tau} = \{{\mathbf{s}}_t,{\mathbf{a}}_t\}_{t=1}^T$, where each state ${\mathbf{s}}_t = \{{\mathbf{o}}_i^t\}_{i=1}^N$ encapsulates the system observations including UAV states and each action ${\mathbf{a}}_t = \{a_i^t\}_{i=1}^N$ denotes periodically inject anomalous signals controlled by classical stress testing methods. Let $\mathcal{P}_{\text{safe}}$ denote the distribution of normal operations and $\mathcal{P}_{\text{crit}}$ represent the distribution of critical failures. Due to its built-in self-healing functionalities, the tested UTM system usually maintains $\mathbb{P}(\boldsymbol{\tau} \in \mathcal{P}_{\text{safe}}) \gg \mathbb{P}(\boldsymbol{\tau} \in \mathcal{P}_{\text{crit}})$.

\subsection{Scenario Definition}
In the context of UTM operations, we define a scenario as a finite slice of system trajectory that captures both the temporal evolution and spatial interactions of managed entities. Formally, a scenario $\xi$ can be represented as a subsequence of state-action pairs within trajectory $\boldsymbol{\tau}$, such that $\xi = \{{\mathbf{s}}_t,{\mathbf{a}}_t\}_{t=k}^{k+\Delta}$, where $k$ denotes the starting time step and $\Delta$ represents the scenario duration.

\subsection{UTM Scenario Discovery: An MDP}

For scenario generation, three main prior approaches exist: policy-based dynamic evolution \citep{okellyScalableEndtoEndAutonomous2019}, iteration-based editing \citep{dingLearningCollideAdaptive2020, liuSafetyCriticalScenarioGeneration2024}, and template-based one-shot generation \citep{dingRealGenRetrievalAugmented2024}. Each approach offers distinct advantages: dynamic evolution methods produce natural interactions but offer limited control, iterative editing provides precise scenario control but may result in mechanical generation processes, while template-based methods trade generation efficiency with diversity. Given the time-invariant nature of UTM systems, the state transitions naturally form a Markov Decision Process (MDP). Transition dynamics between states is dominantly worth studying.

In this framework, the discovery of vulnerability-inducing scenarios can be formulated as an MDP $\mathcal{M} = (\mathcal{S}, \mathcal{A}, \mathcal{P}, r, \gamma)$, where the state space $\mathcal{S}$ captures system configurations, action space $\mathcal{A}$ represents possible perturbations, $\mathcal{P}$ denotes the transition dynamics, $r$ is the reward function, and $\gamma$ is the discount factor.
This formulation, however, presents unique challenges stemming from both UTM system characteristics and technical constraints. First, the centralized nature of UTM systems introduces high-dimensional state spaces that encompass complex interdependencies among multiple UAVs. The system's complexity is further compounded by the need to maintain global consistency while managing local interactions, resulting in a combinatorial explosion of possible scenarios.
Moreover, the technical intrinsics of scenario discovery introduce additional challenges. The distribution of critical scenarios exhibits a pronounced long-tail effect, where vulnerability-inducing states occupy only a small fraction of the state space, i.e., $\mathbb{P}(\mathbf{s} \in \mathcal{S}_{\text{crit}}) \ll \mathbb{P}(\mathbf{s} \in \mathcal{S}_{\text{safe}})$. This distributional imbalance is exacerbated by the lack of expert behavior data in critical scenarios, as most operational data comes from safe system states. Consequently, traditional imitation learning approaches or straightforward policy optimization methods become insufficient for effective scenario discovery.
The transition dynamics $\mathcal{P}$ in this MDP need to capture not only the physical evolution of the system but also the likelihood of transitioning into critical scenarios. Formally, for any state $\mathbf{s}_t$ and action $\mathbf{a}_t$, we define the transition probability as:
\begin{equation}
\mathcal{P}(\mathbf{s}_{t+1}|\mathbf{s}_t, \mathbf{a}_t) = \begin{cases}
p_{\text{safe}}(\mathbf{s}_{t+1}|\mathbf{s}_t, \mathbf{a}_t) & \text{if } \mathbf{s}_{t+1} \in \mathcal{S}_{\text{safe}} \\
p_{\text{crit}}(\mathbf{s}_{t+1}|\mathbf{s}_t, \mathbf{a}_t) & \text{if } \mathbf{s}_{t+1} \in \mathcal{S}_{\text{crit}}
\end{cases}
\end{equation}
where $p_{\text{safe}}$ and $p_{\text{crit}}$ represent the transition dynamics in safe and critical regions respectively.

\subsection{Reinforcement Learning Formulation}
\label{TowardsReinforcementLearning}
In our vulnerability discovery task, scenarios inducing system instability are assigned higher rewards, guiding the exploration towards potential failure modes. Formally, we aim to find a policy $\pi$ that maximizes the expected cumulative reward:
\begin{equation}
J(\pi) = \mathbb{E}_{\tau \sim \pi}\left[\sum_{t=0}^{T} \gamma^t r(\mathbf{s}_t, \mathbf{a}_t)\right]
\end{equation}
where the reward function $r(\mathbf{s}_t, \mathbf{a}_t)$ is designed to favor transitions that push the system towards unstable states, as detailed in following paragraphs.

Given the inherently high-dimensional state space and complex system dynamics of UTM systems, explicit modeling of the state transition probability distribution $\mathcal{P}_a(\mathbf{s},\mathbf{s}')$ faces not only computational scalability challenges, but more fundamentally, modeling complete system dynamics remains largely infeasible \citep{haWorldModels2018}. This limitation is particularly evident in traditional test case design approaches \citep{leeAdaptivestresstesting2015}. Attempts to capture system physics and constraints through rule-based modeling \citep{liuAbstractiondiscretizationrobustness2014} inevitably results in insufficient test coverage or inadequate test quality. As a workaround, we propose a policy-centric alternative that learns action selection strategies rather than directly modeling action prediction, thereby implicitly capturing MDP dynamics in state transitions through strategic exploitation \citep{silvergeneralreinforcementlearning2018}, in order to elegantly circumvents both theoretical and practical challenges of explicit world-modeling.

Specifically, our optimization objective becomes:
\begin{equation}
\max_{\pi} \mathbb{P}(\mathbf{s}_{t+1} \in \mathcal{S}_{\text{crit}} | \mathbf{s}_t, \mathbf{a}_t \sim \pi(\cdot|\mathbf{s}_t))
\end{equation}

\paragraph{State}
The state space of testing framework is a concatenate of temporal, spatial and mission information in UAV numbers. We define $o_i^t \in \mathcal{O} \subset \mathbb{R}^d$ as a vector of $d$ relevant features, which encapsulates the observable information for $i$-th UAV at $t$-th time-step, including:
kinetic information (position, velocity, and acceleration of all UAVs), environmental data (obstacles, weather conditions, and airspace restrictions) and mission-specific details (battery levels, payload capacity, and route destinations).
To capture both cross-UAV dependencies and temporal dependencies, we represent the state $s\in\mathcal{S}$ as a sequence of observations of all $N$ UAVs over a fixed time window $T$, namely, $\mathbf{s} = \{(o_1^1, ..., o_N^1),\dots,(o_1^T,\dots,o_N^T)\}$.

\paragraph{Action}
The action space $\mathcal{A}$ comprises a discrete set of all possible injection operations, each targeting a specific component or aspect of the SUT. We define $\mathcal{A}$ as the Cartesian product of two sets $\mathcal{A} = \mathcal{D}^{'} \times \mathcal{F}$ where $\mathcal{D}^{'}$ represents the set of $N$ targetable UAV, and $\mathcal{F}$ is the set of $m$ applicable disturbance injections. For each injection, all the possible types are listed in Table \ref{action_types_of_policy_model} in Appendix \ref{Sec:App_action_space}.

\paragraph{Reward}
The reward function $r$ is designed to capture the system's safety and operational efficiency as $r(s_t, a_t) = \sum_{i=1}^K \alpha_i r_i(s_t, a_t)$ denoting reward at timestep $t$ where $r_i$ are individual reward components (e.g., collision avoidance, mission completion, system stability) and $\alpha_i$ are their respective weights. Return-to-go is the cumulate sum of reward from current time $t$ to the ending time $T$ as $R_{t:T} = \sum_{i=t}^T r_i$.

\subsection{Choice of Decision Transformer}\label{sec:transformer}

Given UTM's safety-critical nature and biased data distribution discussed earlier, we adopt an offline reinforcement learning approach to discover vulnerability-inducing scenarios. We leverage the Decision Transformer(DT) architecture to effectively model the MDP dynamics, of which the return-to-go design in DT provides a natural guidance.

Self-attention mechanism also provides interleaving data utilization among different head in favor of modeling agent-wise interaction. Our strategy of Transformer usage resides in (1) modeling complex system mechanism through learning reward/return, (2) generating targeted and valuable actions based on knowledge of world. Formally, given a sequence of state-return pairs $(s_1, R_1), ..., (s_T, R_T)$, the decoder-only Transformer processed this information end-to-end into a set of predictions $\{(\hat{R_1}, \hat{a_1}), ..., (\hat{R_T}, \hat{a_T})\}$ with $\hat{R}$ as regressive modeling of world and $\hat{a}$ decision of actions, where $\hat{R_t} = f_\theta(s_1, R_1, ..., s_t)$ and $\hat{a_t} = f_\theta(s_1, R_1, ..., s_t, R_t)$ with $f_\theta$ denoting the Transformer decoding function with parameters $\theta$. By conditioning action generation on desired future returns, we can effectively steer the exploration towards potentially vulnerable states while respecting the constraints of the offline setting. Formally, we learn a sequence prediction model that bridges the distributional gap through return-conditioned generation:

\begin{equation}
p_\theta(\mathbf{a}_t|\mathbf{s}_t, R_{t:T}, \mathcal{H}_t) = \text{Transformer}([\mathbf{x}_{t-l},...,\mathbf{x}_{t}])
\end{equation}

where $\mathbf{s}_{t}$ represents the current system state, $R_{t:T}$ denotes the return-to-go, and $\mathcal{H}_t$ captures the history of previous states and actions. Each token $\mathbf{x}_i$ in the input sequence combines state, action, and return-to-go information through learned embeddings over a context length $l$.

Considering complexity and heterogeneity of UTM run-time states, the transformer's attention mechanism and latent space encoding capabilities significantly enhance exploration efficiency. Self-attention mechanisms capture complex patterns in historical data that correlate with different system outcomes, from safe operations to potential failures. We train the model by optimizing a dual objective that encompasses both action prediction and return estimation:

\begin{equation}
\begin{aligned}
\mathcal{L}(\theta) &= \mathbb{E}_{(\mathbf{s}_t, \mathbf{a}_t, R_{t:T}) \sim \mathcal{D}}[-\log p_{\theta}(\mathbf{a}_t|\mathbf{s}_t,\mathbf{a}_t, R_{t:T}, \mathcal{H}_t)] \\
&+ \mathbb{E}_{(\mathbf{s}_t, \mathbf{a}_t, R_{t:T}) \sim \mathcal{D}}[- \log p_{\theta}(R_{t:T}|\mathbf{s}_t, \mathcal{H}_t)]
\end{aligned}
\end{equation}

To further enhance the efficiency of offline data utilization and address the challenges of imbalanced data distribution and limited expert demonstrations, we augment the transformer architecture with advanced sampling mechanisms. These mechanisms help mitigate the scarcity of expert data and the inherent imbalance in sample distribution, details of which will be elaborated in subsequent sections.

\section{Framework Architecture}

\begin{figure*}[t]
\begin{center}
\includegraphics[width=\textwidth]{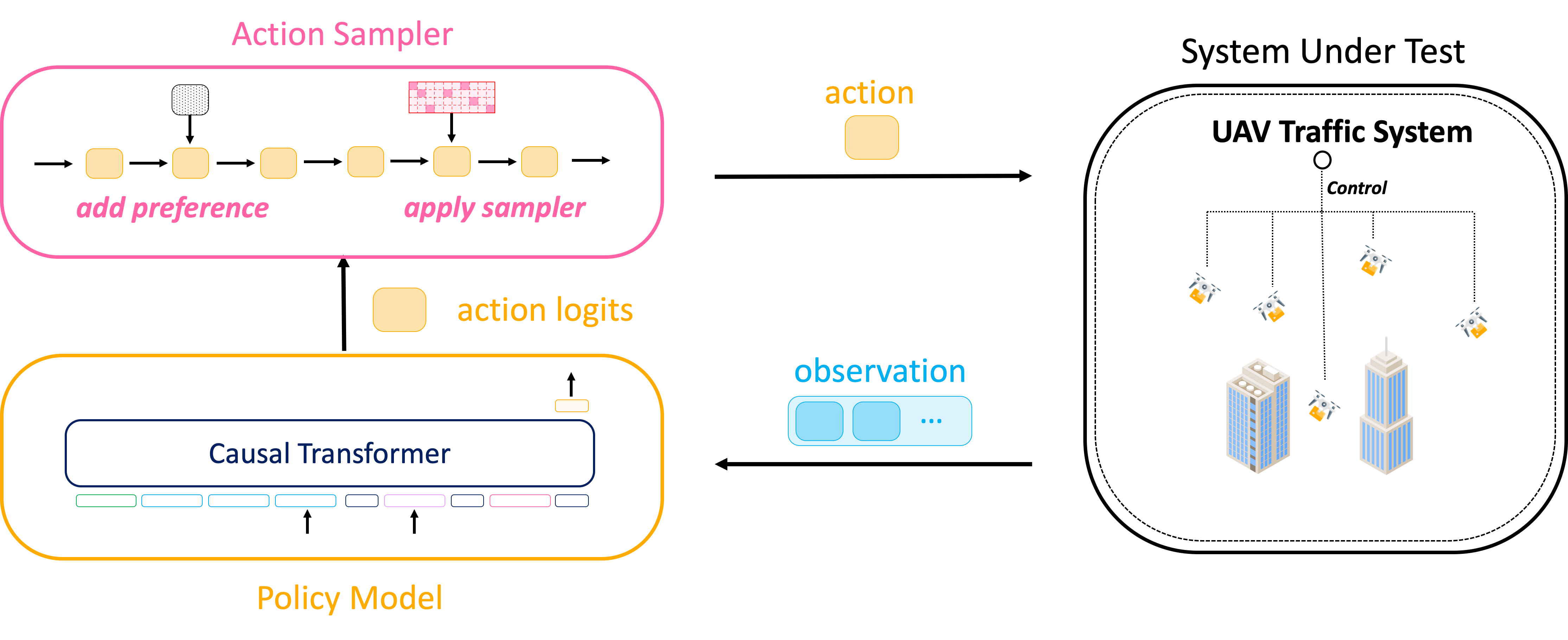}
\end{center}
\caption{\textbf{Architecture overview of the proposed scenario-oriented testing framework.} The framework consists of two primary modules: (1) a Transformer-based Policy Model (PM) for generating fault scenarios based on real-time and historical SUT data, and (2) an Action Sampler (AS) that enforces predefined safety rules and filters out undesirable actions. The validated scenarios are then injected into the System-Under-Test (SUT) for evaluation. This architecture effectively narrows the search space to high-risk scenarios, improving fault detection efficiency and reducing unnecessary exploration of low-risk cases.}
\label{fig:overview}
\end{figure*}

In this section, we present a framework that generates complex testing scenarios while incorporating domain knowledge and expert preferences. As shown in Fig. \ref{fig:overview}, in this framework, we utilize a Transformer-based model as policy model (PM) to generate candidate actions conditioned on system states. According to action space defined in Section \ref{TowardsReinforcementLearning}, generated actions can be interpreted as potential fault injections to be applied to typical victim drones in UTM. Subsequently, these actions are passed through a domain-specific action sampler (AS). AS serves for two purposes: (1) ensure the PM-generated actions available within the specific UTM context; (2) leverage human expert knowledge to re-sample actions with balanced preference bias in chosen actions and agents. Only actions sampled are injected into the system-under-test (SUT). On SUT finishing execution, a new system state would be generated and fed back to the PM, along with the evaluation of the actions (reward). The PM iteratively explores the scenario space to expose potential vulnerabilities.

\subsection{Policy Model}
In this subsection, we describe the design of PM, according to RL formulation defined in \ref{TowardsReinforcementLearning}. The Policy Model serves as the generative engine of our framework, leveraging the power of Transformer architectures to capture complex temporal dependencies and system dynamics. During training, PM serves to model trajectory sequence from UTM and learn internal natures in offline dataset. In performing inference, PM processes real-time UTM context and generates proposed fault injection actions to AS.

\begin{figure}[t]
\begin{center}
\includegraphics[width=\columnwidth]{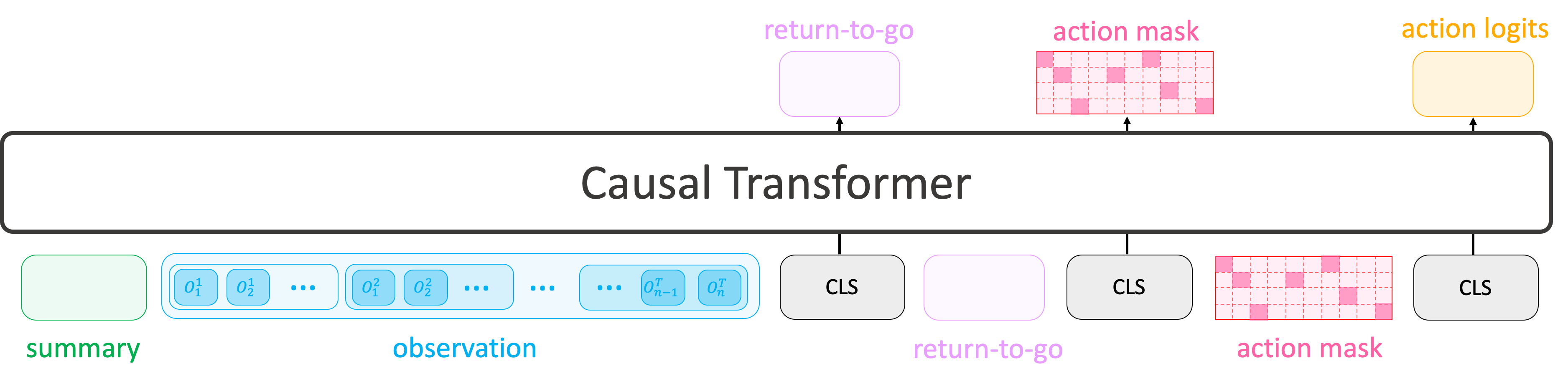}
\end{center}
\caption{\textbf{Architecture of the Policy Model (PM).} The PM utilizes a Transformer-based reinforcement learning framework, taking both historical and real-time SUT states as input tokens to capture temporal dependencies and system dynamics. The model generates action sequences that include both environmental manipulations (e.g., placing obstacles) and internal state changes (e.g., network degradation).}
\label{fig:transformer}
\end{figure}

\paragraph{Time sequence and action modeling}
Expanding on previous work that utilized Transformers for decision-making \citep{chen2021decisiontransformer}, we unify temporal sequences by projecting observations $o$, actions $a$, returns $R$ and rewards $r$ into a homogeneous space. We aggregate rewards $r$ into summary tokens, similar to the construction of return-to-go token $R$ (summarizing $T$ incoming timesteps). Input sequence thus carries data of in-total $3\times T$ timesteps while focusing on central $T$ \textit{current} timesteps. Tokens would then be arranged as array of $\left\langle \mathbf{O},\mathbf{R},\mathbf{A}\right\rangle$ tuples with length of $T$ timesteps. Considering temporal dependency in decision making, we masked out $\mathbf{R}$ and $\mathbf{A}$ tokens except that in last time step. Thus model utilize $T \times N$ observation tokens to predict the current return-to-go token $\hat{R}$ to fit ground-truth return $R$, as learning of implicit system nature. An intermediate \textit{mask} token is introduce to mask out invalid action choices, in favor of modeling system capability according to current state.

\paragraph{Embedding and Causality}
To enhance the modeling of causal dependencies within the policy model, we employ a multi-faceted approach. We augment the sequentially sampled multi-agent drone observation data with positional embedding. Additionally, as shown in Fig. \ref{fig:transformer}, input sequence is augmented with different classification (CLS) tokens as powerful discriminators in order to reduce the ambiguity of prediction targets. Inspired by insights from \citet{shawSelfAttentionRelativePosition2018}, we prioritize the most recent observations by placing them closest to the CLS token, ensuring that the model pays particular attention to the latest information when making decisions. This aligns with the principle that recent events often carry more causal relevance than distant ones.

To  capture long-range dependencies, we employed self-attention mechanism among tokens together with a semi-lower-triangular agent-wise causal mask in attention calculation to preserve decision causality. Observation tokens $o$ at identical timestep are visible to each other homogeneously. However the \textit{$\hat{R}$} tokens could be predicted with only \textit{observation} tokens visible before being fed with ground-truth \textit{return-to-go} token. And only older $\left\langle \mathbf{O},\mathbf{R},\mathbf{A}\right\rangle$ tuples are visible to newer ones. We aim to guide the model to construct a more comprehensive and nuanced understanding of the causal dynamics. Formally, we can sequentially express the prediction task as $\hat{a}_t = f_\theta(S_{-t:1}, o_{1:t}, R_{t}, M_{t})$ where $S$ denotes the \textit{summary} token aggregating previous $T$ time steps and $f_\theta$ represents the Transformer model with parameters $\theta$.

\subsection{Action Sampler}
Inductive bias and generality are key drawbacks of traditional offline RL methods. We design a set of sampling strategies as a workaround. In this subsection, We first introduce preference bias as a notation of human feedback. And we describe action sampler functions between PM and SUT.

\begin{figure}[t]
\begin{center}
\includegraphics[width=\columnwidth]{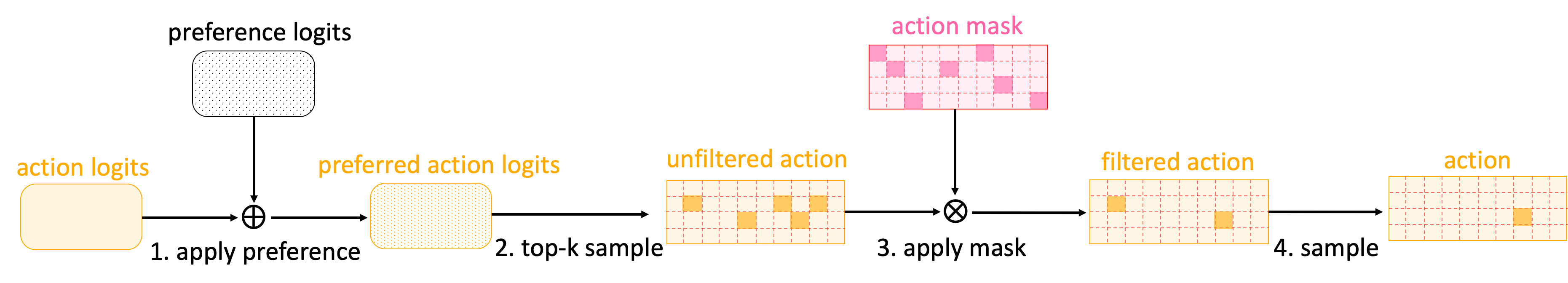}
\end{center}
\caption{\textbf{Pipeline of the Action Sampler (AS).} The AS enforces safety constraints and domain-specific rules, filtering out irrelevant actions generated by the Policy Model (PM) before injecting them into the System-Under-Test (SUT), ensuring the integrity of the testing process.}
\label{fig:action_filter}
\end{figure}

\paragraph{Preference Bias}
In offline RL with autoregressive models, the collected training data often exhibits biased distributions. This limits the sampling of rare events. Meanwhile, the more complex the system is tested, the more insidious the vulnerability and the more significant the long-tail effect. In this work, training dataset is collected through traditional stress testing, where unpredictable inductive bias is common in production systems.

We introduce \textit{Preference Bias}, improved from popularity bias \citep{Klimashevskaia_2024} with additional domain expert knowledge, to unify imbalance in model prediction and gap in prior human preference. Preference bias carries a expected distribution of $\left \langle \mathrm{UAV}, \mathrm{Action} \right \rangle$ tuples. The output of the offline-trained PM is augmented with compensation dynamically calculated from distance between recent historical trajectories and given distribution.

\paragraph{Action Candidate Sampling}
As shown in Fig. \ref{fig:action_filter}, we adjust PM's predicted action logits using the preference distribution. To address long-tail effect and improve fairness \citep{menonLongtaillearninglogit2020}, Top-K sampling is introduced after augmentation in order to maintain variance. To maintain physical feasibility, immediate action mask is applied in order to filter intolerable action candidates. The final action is sampled through a uniform sampling after masking.

\section{Evaluation}
We train the proposed framework with a large-scale offline dataset of around 17B tokens collected from stress testing data and evaluate on an industry-level simulator.
As is summarised in Table. \ref{tab:region} in Appendix \ref{sec:envrionment_details}, the training set consists of seven distinct regions and online testing includes two regions.
The training dataset covering diverse geographical and operational characteristics, including a mix of rural (12.2\%), suburban (39.0\%), and urban areas (48.8\%), each with varying numbers of UAVs, airports, and flight lines. The dataset is balanced to represent the typical distribution of scenarios encountered in real-world UTM systems.
For testing, two regions (TR1 and TR2) are excluded from the training set to provide evaluations of the generalization capabilities.

We design two model of different size, with 1.2 billion and 2 billion parameters (referred as PM-1.2B and PM-2B respectively). We train each model on 16 NVIDIA A100 GPUs, each equipped with 80GB of memory. The training utilized PyTorch's Distributed Data Parallel (DDP) to efficiently distribute the workload across multiple GPUs, ensuring high computational efficiency and resource utilization. During training, the dataset is divided into smaller slices of 3B tokens for sequential loading during training.

\begin{figure*}[t]
    \centering
    \includegraphics[width=0.5\linewidth]{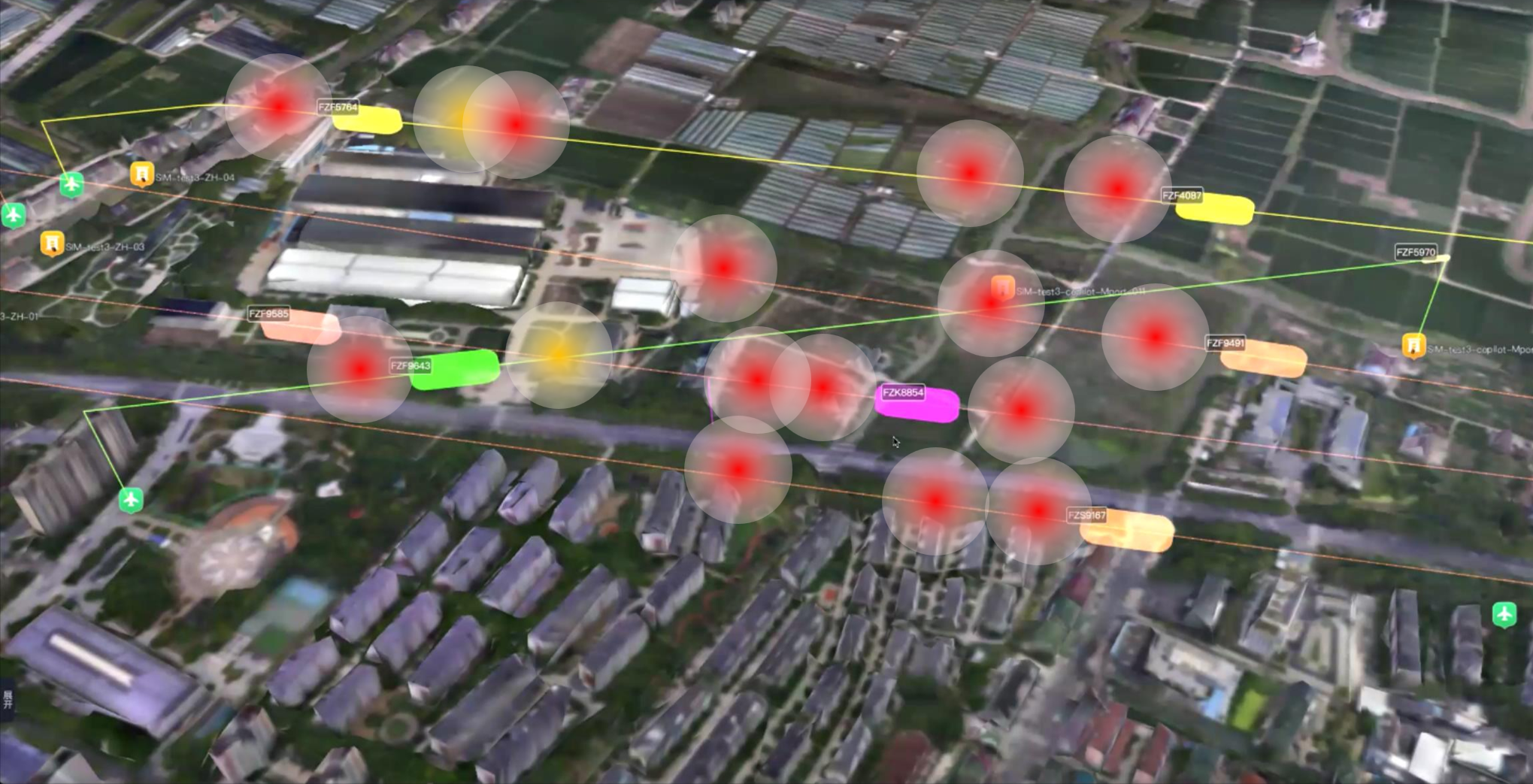}
    \caption{Examples of detected UTM fault scenarios, where \textcolor{yellow}{yellow} marks indicate fault cases revealed by traditional expert method and \textcolor{red}{red} marks indicate those revealed by the framework.}
    \label{fig:UTM_fault_demo}
\end{figure*}

We evaluate the performance of the proposed model through both offline and online evaluations to provide a comprehensive analysis.
In Section \ref{sec:offline_eva}, we focus on the offline evaluation of the PM's behavior during training, where we analyze the evolution of action accuracy and return-to-go loss.
In Section \ref{sec:online_eva}, the online evaluation measures the model's performance in a deployed real-world environment, where we collect and analyze a range of key metrics. This dual evaluation framework offers a holistic view of the model's efficacy, ensuring robustness both during training and in practical applications. As shown in Fig. \ref{fig:UTM_fault_demo}, our framework is able to reveal much more potential vulnerabilities than conventional hand-crafted strategy does.

\subsection{Offline Evaluation}
\label{sec:offline_eva}

For offline evaluation, we focus on the impact of model size on action accuracy and return-to-go loss during training.
Especially, we apply the top K action accuracy in that in our framework, actions are sampled based on the top-k predictions rather than solely the top-1. We further detail the impact of top-1 accuracy in Appendix.\ref{Appendix.UTM}.
The results in Fig. \ref{fig:offline_eval} illustrate that larger models consistently perform better across both action accuracy (highest) and return-to-go loss (lowest) metrics. This indicates that larger models have a better capacity to capture the underlying structure in the offline data, achieving more accurate action selections with fewer training tokens.
Fig. \ref{fig:offline_eval} also reveals that the PM-2B model begins to overfit much later compared to the smaller PM-10M and PM-100M models.
This suggests that larger models not only perform better in terms of action accuracy but also exhibit better generalization properties, allowing them to continue learning effectively with more data before encountering overfitting issues. This behavior is a hallmark of the scaling effect, where larger models benefit from increased capacity and more robust training dynamics, making them more resistant to overfitting compared to smaller models.

\begin{figure*}[ht]
    \centering
    \begin{subfigure}[t]{0.3\textwidth}
        \begin{center}
        \includegraphics[width=\textwidth]{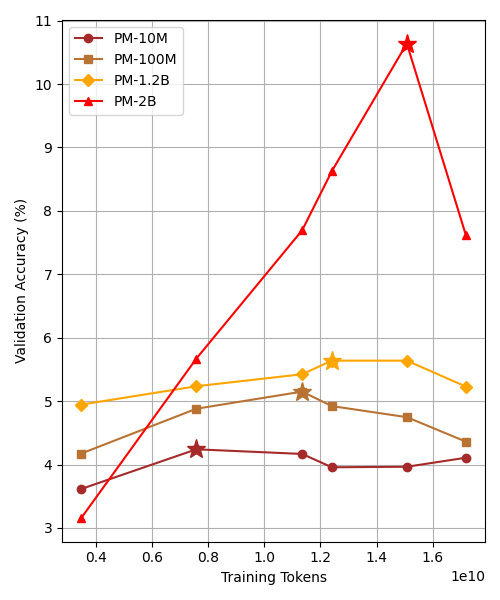}
        \end{center}
        \vspace{-5pt}
        \caption{Action accuracy of PM;}
        \label{fig:action_acc}
    \end{subfigure}
    \begin{subfigure}[t]{0.3\textwidth}
        \begin{center}
        \includegraphics[width=\textwidth]{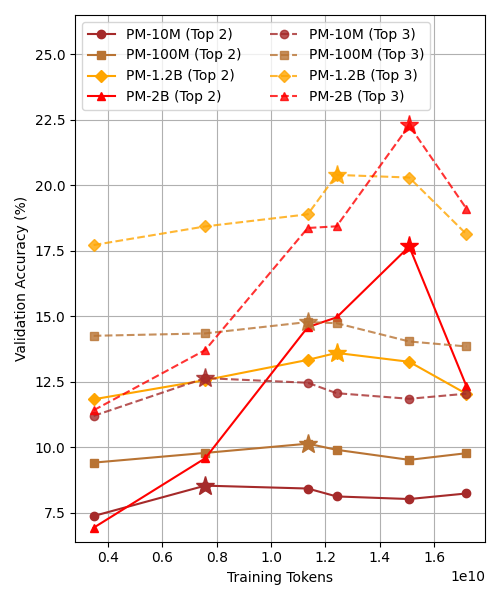}
        \end{center}
        \vspace{-5pt}
        \caption{Top 2/3 action accuracy;}
        \label{fig:action_acc_top_2/3}
    \end{subfigure}
    \begin{subfigure}[t]{0.3\textwidth}
        \begin{center}
        \includegraphics[width=\textwidth]{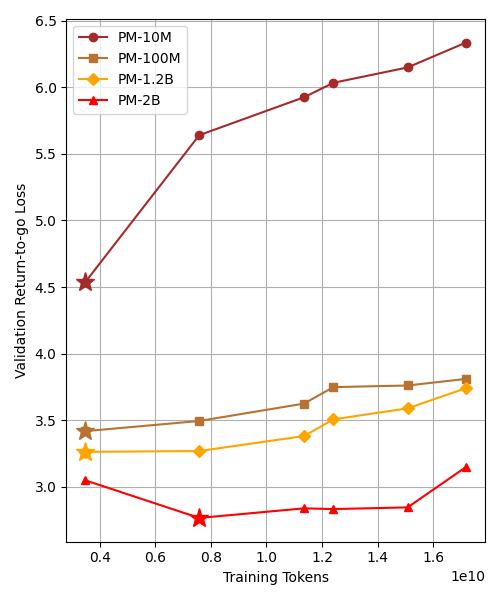}
        \end{center}
        \vspace{-5pt}
        \caption{Return-to-go loss of PM;}
        \label{fig:rtg_loss}
    \end{subfigure}
    \caption{\textbf{Offline evaluation results on validation sets during training.} The action accuracy and return-to-go of the models (PM-10M, PM-100M, and PM-2B) measured over increasing training tokens on validation sets. All models show an initial increase in accuracy, followed by a decline, indicating overfitting phenomenon. Similarly, all models eventually increase in return-to-go loss, signaling overfitting. Larger models demonstrate a clear advantage, achieving significantly higher accuracy lower return-to-go loss compared to the smaller models. The peak action accuracy for each curve is highlighted with a star.}
    \label{fig:offline_eval}
\end{figure*}

\subsection{Online Evaluation}
\label{sec:online_eva}
To evaluate the effectiveness of proposed framework in unseen environments, we selecte several key metrics to evaluate the preference and effectiveness of PM, as well as the quality of actions, as is shown in Table. \ref{tab:metric}. For detailed explanation of each metric, we refer to the Appendix \ref{sec:metric_appendix}.

\renewcommand{\thefootnote}{\fnsymbol{footnote}}
\begin{table}[t]
    \centering
    \resizebox{\columnwidth}{!}{
    \begin{tabular}{ccccccccc}
    \toprule
    \multirow{2}*{Metrics}  &  \multicolumn{2}{c}{PM-2B} & \multicolumn{2}{c}{PM-1.2B} & \multicolumn{2}{c}{Expert-Guided Exploitation} & \multicolumn{2}{c}{Smoke Test\footnotemark[1]} \\
    \cline{2-9}
                            &  TR1 & TR2       & TR1 & TR2       & TR1 & TR2                          & TR1 & TR2 \\
    \midrule
    APO(\%)     & \textbf{20.0} & \textbf{31.5} & 55.3 & 38.3 & 72.0 & 83.3 & 100 & 100\\
    APD(\%)     & 26/34/21/19 & 46/32/11/11 & 28/27/22/23 & 30/29/20/21 & \textbf{25/25/25/25} & \textbf{25/25/25/25} & N/A & N/A \\
    \midrule
    HAR(\%) & \textbf{10.8} & \textbf{4.9} & 6.7 & 4.2 & 3.6 & 1.7 & N/A & N/A \\
    CAR(\%) & \textbf{29.7} & \textbf{64.1} & 4.0 & 4.5 & 4.1 & 3.9 & N/A & N/A \\
    \midrule
    SPM & \multicolumn{2}{c}{\textbf{50.5}} & \multicolumn{2}{c}{17.6} & \multicolumn{2}{c}{5.8} & \multicolumn{2}{c}{N/A} \\
    FPM & \multicolumn{2}{c}{\textbf{7.6}}  & \multicolumn{2}{c}{2.2} & \multicolumn{2}{c}{$<$1.0\footnotemark[7]} & \multicolumn{2}{c}{$<$1.0\footnotemark[7]} \\
    \bottomrule
    \end{tabular}}
    \caption{\textbf{Performance metrics of the propose framework in online environments of unseen regions.} This table shows the online results in out-of-distribution region TR1 and TR2. Results of PM models are reported on over 700 hours testing in total, with around 100M records for each model in each region. The detailed definition of metrics can be found in Table. \ref{tab:metric} and Appendix \ref{sec:metric_appendix}.
    }
    \label{tab:main_exp}
\end{table}

\footnotetext[1]{The smoke testing refers to the basic functionality testing of UTM system. This is conducted as the initial testing after a new build or version of the UTM system.\label{footnote:smoke_test}}
\footnotetext[7]{The FPMs are below 1.0 because the two baseline tests have already been thoroughly used to identify existing bugs and improve UTM in advance, while our method is focused on discovering new bugs in the updated version of the UTM system after the baselines have reached their detection limits.\label{footnote:FPM}}
\renewcommand{\thefootnote}{\arabic{footnote}}

From the results shown in Table. \ref{tab:main_exp}, we can conclude that the proposed PM-2B model significantly outperformed both expert-guided testing and smoke test baselines across all key metrics. Specifically, PM-2B generates high-risk scenarios weight times faster than smoke testing, and is able to discover bugs while expert-guided testing method fails to. This indicates that the proposed framework is more effective in identifying critical scenarios and potential failures.
Furthermore, comparing with smaller PM-1.2B model, PM-2B performs significantly better in action quality and efficiency.
This suggests the existence of scaling effect between model size and online performance in discovering critical cases and efficiently covering high-risk regions.
Interestingly, the PM-2B model detected failure modes (SPM and FPM) that the smoke test completely missed. This emergent capability shows that the PM framework can find faults beyond traditional rule-based methods, demonstrating its utility for uncovering rare bugs.
Considering both the scaling effect and emergent abilities, our framework shows significant promise for scaling up model sizes, and has the potential to become a breakthrough in the testing field in the future.
However, PM models fail to balance the distribution of different action types, which could lead to potential under-exploration in less frequent action spaces.
This suggests a need for better action sampling strategies.

\section{Discussion}
\paragraph{Why don't we use inverse RL?}
One might consider using inverse reinforcement learning (IRL) to infer the underlying reward function from the system's operational trajectories and then optimize for its opposite to discover vulnerabilities. However, IRL faces several fundamental limitations in UTM testing. The primary issue lies in the inherent ambiguity of the observed system behaviors. Unlike traditional IRL settings where expert demonstrations represent optimal or near-optimal policies, our historical trajectories mostly consist of operations that finally result in ``safe'' states, making it difficult to reliably invert the system's true safety objectives. Furthermore, there exists a fundamental tension in the learning objectives: strictly imitating the suboptimal aspects of historical operations might perpetuate existing blindspots in testing coverage, while over-idealizing the system's intended behavior could lead to unrealistic vulnerability scenarios. This inherent ambiguity, combined with the need for active failure exploration, makes IRL less suitable than our direct policy learning approach.

\paragraph{Why does proposed framework exceed the performance of human experts?}
Although trained with expert-guided exploitation data, PM model ultimately surpass the performance of human experts.
This is attributed to that PM model applies offline RL, which can be viewed as an implicit filter of low-quality actions \citep{prudencio2023survey}, making it less susceptible to distraction during the search for long-tail scenarios.

We can illustrate this by analyzing the hazard action ratio per observation, which is obtained by multiplying HAR and APO, and the constant-pressure action ratio per observation, calculated by multiplying CAR and APO. For both PM-2B, PM-1.2B, and human experts, the hazard action ratio per observation is consistently around 2\%. This shows that all methods are similarly effective in identifying high-risk actions.
However, the key difference is that the PM models demonstrate a significantly higher constant-pressure action ratio per observation, indicating that they maintain a more sustained level of high-risk actions over time. This ability to constantly pose challenges and maintain pressure highlights the advantage of the PM models in exploring complex, high-risk scenarios more thoroughly, thereby leading to superior fault detection and scenario coverage.

\section{Conclusion}
We propose a novel scenario-oriented testing framework for vulnerability detection in mission-critical systems, specifically applied to UTM. Our approach leverages a Transformer-based policy model to tackle long-tail effect and efficiency challenge in fault detection. Context utilization in policy model improves generality in unseen regions.
Our results highlight the potential of learning and expert hybrid approaches in fortifying mission-critical systems. This work opens new directions for end-to-end auto-regressive learning in safety-critical system testing. Future work could explore the application of this framework to other mission-critical domains beyond UTM, such as autonomous vehicles or industrial control systems.

\section*{Impact Statement}

This work aims to enhance the safety and reliability of Unmanned Traffic Management systems, which have significant societal implications as aerial mobility becomes increasingly important in urban environments. While our framework's ability to discover system vulnerabilities advances testing capabilities, we acknowledge that such knowledge could potentially be misused, and have therefore designed our methodology to be accessible only to authorized system developers and testers, with appropriate safeguards for responsible disclosure. The improved efficiency in detecting critical scenarios could accelerate UTM deployment, potentially affecting traditional air traffic management roles while creating new opportunities in system development and maintenance. Furthermore, our approach may have broader applications in other safety-critical domains such as autonomous vehicles, medical devices, and industrial control systems, underscoring the importance of maintaining rigorous ethical standards and careful consideration of failure consequences as the technology is adapted to new contexts. We are committed to open dialogue with stakeholders and the research community about these implications as the technology continues to evolve.


\bibliographystyle{unsrtnat}
\bibliography{references}

\appendix

\section{UTM System Architecture and Testing Pipeline}
\label{Appendix.UTM}

\paragraph{What is Unmanned aircraft Traffic Management (UTM) system?}
The Unmanned aircraft Traffic Management (UTM) system, as introduced by the National Aeronautics and Space Administration (NASA) \citep{kopardekar2014unmanned,kopardekar2016unmanned}, is designed to ensure safe and efficient operation of multiple unmanned Unmanned Aerial Vehicles (UAVs) in shared airspace. The UTM concept is developed to support the integration of UAVs into airspace without requiring human air traffic controllers to manage every UAV directly. Instead, UTM emphasizes the use of automated systems to coordinate UAV operations. This includes services like geofencing, route optimization, and deconfliction, ensuring that UAVs can safely and autonomously operate in both sparsely populated rural and densely populated urban areas or alongside manned aircraft.

UTM is typically developed as a complex system. This is because the UTM systems should integrate a wide range of functionalities and address diverse challenges associated with managing UAV operations in dynamic and unpredictable environments. UTM systems need to handle real-time communication between UAVs, ground stations, and other stakeholders, while simultaneously ensuring safety, efficiency, and fairness in airspace usage.

\begin{figure}[ht]
\begin{center}
\includegraphics[width=0.8\textwidth]{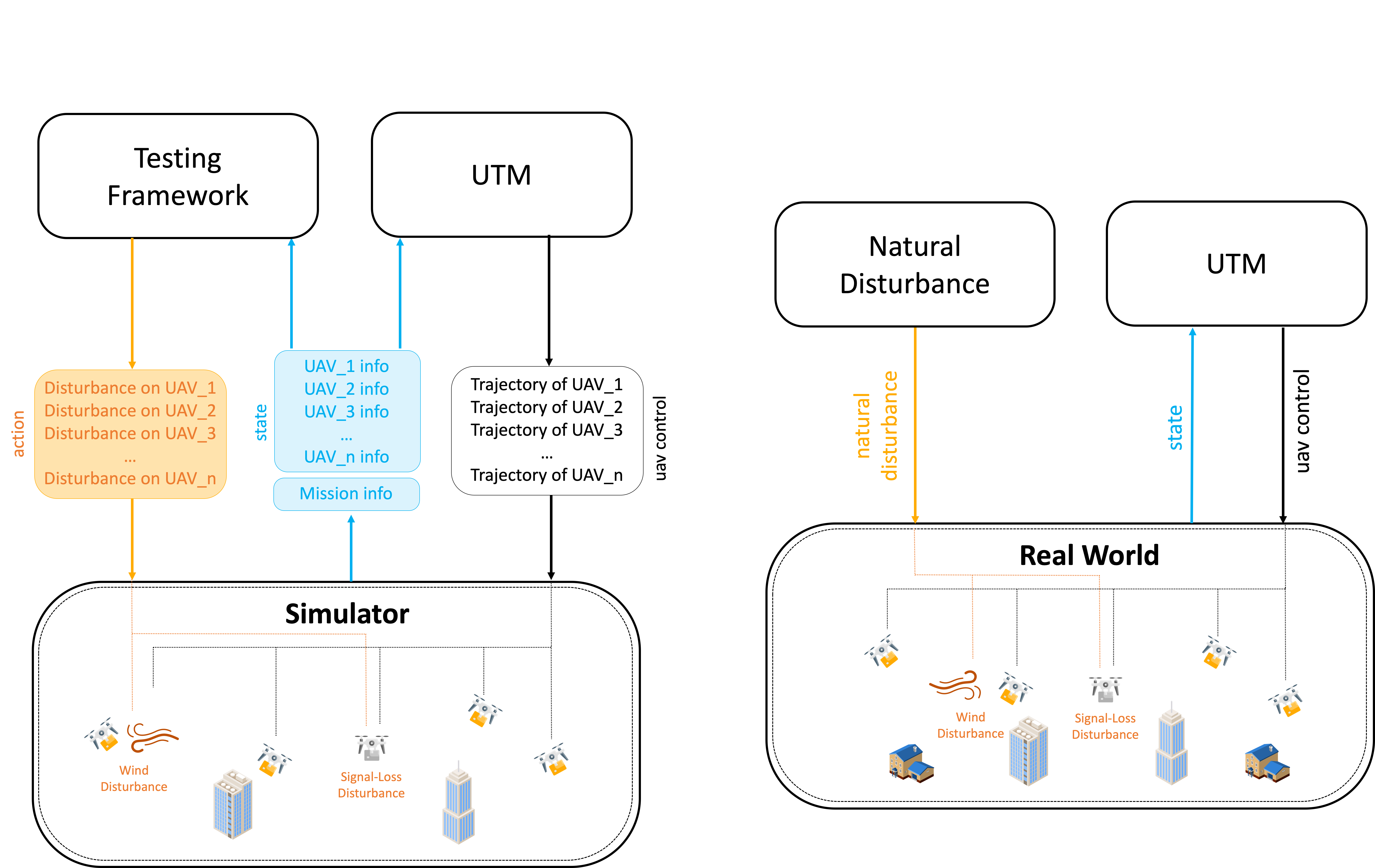}
\end{center}
\caption{\textbf{UTM System and Testing Framework Architecture.} The testing framework works as copilot of UTM and operates on the server-side. As a mission critical system, UTM under test is designed as centralized architecture at once to insure the safety and remove potential conflicts in advance \citep{spalas2024towards,hamissi2023survey}. To align with the design of UTM, our proposed testing framework is also designed centrally. The testing framework mimics the natural disturbance to generate different scenarios.}
\label{fig:UTM_system}
\end{figure}

As show in Figure \ref{fig:UTM_system}, UTM serves as the central coordinator, processing dynamic information received from all UAVs and managing overall traffic flow through sophisticated decision-making algorithms simultaneously. UTM maintains continuous communication, flight route allocation and trajectory assignment with multiple UAVs, each equipped with various sensors and control systems, while simultaneously monitoring environmental conditions and potential conflicts.

\paragraph{What is fault detection in development of UTM and why it is important?}
We define the term \textit{fault detection} as the process identifying possible faults in the UTM system during testing phase, which is before the UTM system is deployed in real-world environments. It is typically divided into several steps, including module testing, integration testing, smoke testing (functional testing), stress testing, etc. After each testing step, the confidence (e.g., reliability, fault tolerance, and compliance with regulatory standards) of UTM system increases as potential faults are identified and addressed, ensuring that the system becomes progressively more robust and reliable.

Fault detection is a critical aspect of UTM development because it directly impacts the safety, reliability, and efficiency of development pipeline. As a mission critical system, the UTM system should be designed to eliminate all the faults it may occur, which are usually costly or even deadly (e.g., UAV crushes, collisions with buildings or even collisions with human injuries) \citep{kopardekar2014unmanned,kopardekar2016unmanned}. By identifying and addressing potential faults during the testing phase, fault detection ensures that the UTM system operates as intended, mitigating risks before deployment in real-world environments. This proactive approach prevents costly failures, enhances system robustness, and builds trust among stakeholders.

\paragraph{Why fault detection is challenging?}

Fault detection in UTM systems is inherently challenging, particularly as testing progresses through advanced stages. While early testing steps may uncover obvious issues, the long-tail of rare and hard-to-detect faults often remains persistent and elusive. This difficulty is compounded by the self-healing capabilities of modern UTM systems, which can mask subtle issues that may only emerge under specific conditions. As is listed in the Table \ref{tab:long_tail_fault}, although several testing steps have been conducted, there still remains faults to threat the safety of the UTM system (e.g. shakedown effects found by Federal Aviation Administration in field testing) \citep{rios2017utm,FAA_2023}. Based on the stepwise testing and field testing results, we estimate the faults found in different steps of testing, as listed in Table \ref{tab:long_tail_fault}. From data in the table, we can see that as several testing steps are conducted, there still exists faults to be detected, which is fatal in mission critical systems.

\paragraph{Why lower top-1 accuracy doesn't impact framework reliability}
While the top-1 action accuracy of our framework may appear relatively low (around 20\% for PM-2B in TR1), this does not compromise the framework's reliability in vulnerability detection for several reasons. First, unlike traditional classification tasks where precise prediction is crucial, our framework operates in the context of safety testing where the primary goal is to discover diverse failure scenarios rather than to predict specific actions. The lower top-1 accuracy actually reflects the framework's ability to explore a broader range of potential actions rather than being constrained to the most probable ones. This is evidenced by our high HAR (10.8\%) and CAR (29.7\%) metrics, which indicate that a significant portion of the generated actions effectively probe system vulnerabilities. Second, our framework employs top-k sampling strategy during testing, which allows it to maintain a balance between exploration and exploitation. The high top-3 accuracy (shown in Fig. \ref{fig:offline_eval}(b)) demonstrates that while the model may not always select the optimal action first, it consistently maintains valuable action candidates within its top choices. Furthermore, when examining the detected vulnerabilities (as shown in Table \ref{tab:main_exp}), our framework achieves significantly higher SPM (50.5) and FPM (7.6) compared to expert-guided testing (5.8 and <1.0 respectively), indicating that the framework's action selection strategy, despite lower top-1 accuracy, is more effective at uncovering critical system faults. This aligns with recent findings in reinforcement learning literature that suggests maintaining action diversity can be more valuable than maximizing prediction accuracy when exploring rare but critical events in the state space.

\begin{table}[htbp]
    \centering
    \resizebox{\textwidth}{!}{\begin{tabular}{cccccc}
    \toprule
     \textbf{Fault Types} & \textbf{Module Testing} & \textbf{Integration Testing} & \textbf{Smoke Testing} & \textbf{Stress Testing} & \textbf{Fault Remaining} \\
    \midrule
     Module Level  & $\sim 20\%$ & $\sim 10\%$ & $\sim 30\%$ & $\sim 40\%$ & $\sim 0.1\%$ \\
     Interface Level  & $\sim 10\%$ & $\sim 20\%$ & $\sim 30\%$ & $\sim 40\%$ & $\sim 0.1\%$ \\
     Running time  & $\sim 10\%$ & $\sim 10\%$ & $\sim 40\%$ & $\sim 40\%$ & $\sim 0.1\%$ \\
     \midrule
     \textbf{Scenario Complexity} & Simple & Simple & Medium & Medium & High \\
    \bottomrule
    \end{tabular}}
    \caption{\textbf{Fault Types Detection during Different Steps of Testing.} The module testing verifies individual components of UTM to ensure they function correctly in isolation. The integration testing checks interactions between combined modules to detect interface issues. The smoke testing ensures basic functionality works correctly after a new build or update, acting as a preliminary check. The stress testing evaluates system stability and performance under extreme or peak load conditions. The tested scenarios for moduel testing and integration testing are relatively simple, while smoke testing and stress testing will generate more complex testing scenarios. As the testing steps conducted one by one, the software maturity of UTM increases gradually. However, there still exists rare faults happening in complex scenarios.}
    \label{tab:long_tail_fault}
\end{table}

\section{Proposed Testing Framework}

\paragraph{Testing Framework}
Testing framework introduced in this work serves as a copilot with UTM, rather than deploying on individual UAV. It monitors identical data streams along with UTM, including UAV telemetry (position, velocity, mission status) and system state information. The UTM system provides trajectory schedule in favor of system robustness, while testing system generating adversarial disturbance actions to increase systematic vulnerability.

Testing system is designed to manipulate external disturbances to UAVs like wind, obstacle and network jitter as shown in Table \ref{action_types_of_policy_model}. Internal functionality and and robustness of on-device system of individual UAV is out of the scope of this research.

\paragraph{Sim vs Real}
The framework's methodology emphasizes systematic exploration of edge cases and rare failure modes that might otherwise remain undiscovered in conventional testing approaches. Environmental disturbances suffer from randomness and difficulty in interpreting. In this work, we make use of simulators which enables configurable environmental disturbances and concrete mapping between them and consequential operating status, in favor of typical analysis and diagnosis. Visibility and capability of UTMs are strictly aligned in whether simulated or realistic context.

Besides, precise timing selection of disturbance injections is within consideration as well. Traffic pressure of UTM for complicated UAV MASs varies with time. Testing system learns to inject actions when UTM is handling the most vulnerable cases in favor of significance of tesing scenarios generated.

\section{Challenges of Testing UTM}

\paragraph{Critical Fault Distribution Imbalance}
While UTM's fault-tolerant design successfully handles most anomalies through automated recovery mechanisms and redundant control strategies, this architectural resilience paradoxically increases the complexity of identifying severe failure scenarios, as intermediate failure states are often automatically corrected before they can develop into observable system failures. Critical failures, those capable of overwhelming the system's self-healing mechanisms, occupy an extremely small portion of the state-action space, which often reside in narrowly defined regions of the state-action space, requiring precise combinations of multiple adverse factors to overcome the system's multi-layer safety functionality. These regions are characterized by specific configurations of multiple elements: particular spatial arrangements of UAVs, precise timing of control actions, specific environmental conditions. Furthermore, these failure scenarios often represent emergent behaviors arising from subtle interactions between multiple system components and their recovery attempts, rather than simple violations of individual safety constraints.

\begin{table}[htbp]
    \centering
    \resizebox{\textwidth}{!}{
    \begin{tabular}{cccccc}
    \toprule
        Types & \makecell{Number of \\Influenced UAVs} & \makecell{Disturbance Times\\ within 60s} & Case Example & \makecell{Real-World \\ Ratio} & Complexity\\
    \midrule
        Safe Flight & 0 & 0 & N/A & $\sim$ 94\% & Low \\
        \midrule
        \multirow{4}{*}{Disturbances} & 1 & 1 & Winds with exceeding magnitude & $\sim$ 5\% & Medium \\
        & $\geq 2 $ & 1 (each) & Winds hit multiple UAVs& $\sim$ 1\% & Medium \\
        & 1 & $\geq 2 $ & Winds hit twice with 60s interval& $\sim$ 0.1\% & High \\
        & 1 & $\geq 2$ (simultaneously) & Signal Loss when Winds hit& $\sim$ 0.01\% & High\\
    \bottomrule
    \end{tabular}}
    \caption{\textbf{Real-World UTM failure distribution.} In real-world UAV fleets, advanced UTM provides fundamental guarantee for safe flight, where faults with increasing risk still exist at a relatively low ratio and are increasingly hard to locate and tackle.}
    \label{tab:Real_World_UTM_failure_distribution}
\end{table}

\paragraph{High-Dimensional State-Action Temporal Dependency}
Testing of UTM systems confronts a fundamental challenge in navigating its inherent high-dimensional state-action coupling relationships. The state space encompasses multiple critical dimensions: spatial coordinates and velocity vectors of each UAV, environmental conditions, and communication network states. Each additional UAV exponentially expands this state space, creating a combinatorial explosion in the dimensions that must be considered during testing. Unlike traditional control systems where failures often manifest through immediate state violations, UTM system failures additionally emerge from specific combinations of historical state sequences and multi-agent coupling, as shown in Table \ref{tab:Real_World_UTM_failure_distribution}. The behavioral trajectory of each UAV is intrinsically influenced by both its historical states and the temporal evolution of other agents' states in the shared airspace. For instance, a seemingly safe trajectory adjustment by one UAV could create cascading effects leading to system-wide conflicts minutes later through complex agent interactions. Furthermore, subtle perturbations in early states can propagate through the system's temporal dynamics to trigger critical failures in significantly later stages. The challenge is particularly pronounced in scenarios involving dense multi-UAV operations, where system behavior emerges from the intricate interplay of multiple agents' temporal trajectories rather than simple state-transition patterns.

\section{Motivation for Transformer and Comparison with Other Models}

The main motivation of applying Transformer as backbone model lies in that the Transformer models are proved to be scalable in multi tasks (e.g., natural language processing \citep{brown2020language}, computational vision \citep{pan2021scalable}, robotics \citep{chebotar2023q},  etc.). The scalability is of essential importance in the development of testing framework in that (1) complex temporal and inter-agent dependencies with scalable sizes of UAV swarm and temporal context window, and (2) long-tail effect in fault distribution requiring sufficiently large dataset to identify faults and to feed in backbone models. Leveraging the Transformer's inherent scalability in modeling extended context lengths and processing large-scale data inputs, it can effectively model complex temporal sequences and inter-agent interactions within UAV swarms of varying sizes. This capability allows the testing framework to accommodate extensive datasets necessary for identifying rare faults due to the long-tail effect in fault distribution. Furthermore, the Transformer's ability to handle large-scale data inputs ensures that the model remains robust and accurate as the system under test evolves (e.g. different region settings, as demonstrated in Table \ref{tab:online_ood}). Consequently, integrating the Transformer as the backbone model enhances the framework's capacity to detect, analyze, and predict system behaviors across diverse operational scenarios.

However, alternative backbone models such as Graph Neural Networks (GNNs), Recurrent Neural Networks (RNNs), Long Short-Term Memory networks (LSTMs), and online reinforcement learning algorithms like Deep Q-Networks (DQNs) or Proximal Policy Optimization (PPO) often struggle to address aforementioned challenges effectively. These models may lack the inherent ability to capture long-range dependencies or scale efficiently with increasing sequence lengths and swarm sizes. Specifically,
\begin{itemize}
    \item RNN/LSTM: RNNs and LSTMs encounter difficulties when modeling long temporal contexts due to issues like vanishing gradients, which add to the training difficulty. What's more, RNNs and LSTMs are hard to parallelized, which adds to the training time, especially when deal with large datasets \citep{devlin2018bert}. Base on our primely experiments, we find that for models below 10 million parameters, RNNs are 10 times slower than Transformers, which constrains the scalability of RNNs.
    \item GNN: GNNs may not scale well with large and dynamic swarm networks, especially when temporal dynamics are involved.
    \item DQN/PPO: DQN and PPO require extensive online exploration and interactions \citep{levine2020offline}, making them less practical for fault detection in complex systems with long-tail fault distributions.
\end{itemize}

\section{Online Evaluation of Out-of-distribution and In-distribution Dataset}

\begin{table}[ht]
    \centering
    \begin{tabular}{ccccc}
    \toprule
    Test Region & APO (\%) & APD (\%) &HAR (\%) & CAR (\%) \\
    \midrule
    TR1 (OOD) & 20.0 & 26/34/21/19 & 10.8 & 29.7 \\
    TR2 (OOD) & 31.5 & 46/32/11/11 & 4.9 & 64.1 \\
    \midrule
    R4 (ID) & 27.3 & 16/29/29/26 & 6.5 & 48.7 \\
    \bottomrule
    \end{tabular}
    \caption{\textbf{Performance metrics of PM-2B .} The metrics include Action Probability per Observation (APO), Action Probability Distribution (APD), High-Value Action Ratio (HAR), and Constant-Pressure Action Ratio (CAR). Testing was conducted in three distinct regions: TR1 (rural, out-of-distribution), TR2 (urban, out-of-distribution), and R4 (suburban, in-distribution), to evaluate the model's generalization capability across diverse environments.}
    \label{tab:online_ood}
\end{table}

As is illustrated in Table.~\ref{tab:online_ood}, the PM-2B model demonstrates strong generalization across different environments, maintaining high performance in both in-distribution (ID) and out-of-distribution (OOD) regions. In the OOD rural region (TR1 \& TR2), the model achieves the comparable performance with ID region (in the context of comparing APO, HAR, and CAR). In contrast, the model's performance in the ID region (R4) shows more balanced APD values (16/29/29/26) than in OOD region, which could be a signal of overfitting.

\section{Online Evaluation Metric Details}
\label{sec:metric_appendix}
In this section, we provide a detailed explanation to selected metrics listed in Table. \ref{tab:metric}.
\begin{table}[ht]
    \centering
    {
    \begin{tabular}{ccc}
    \toprule
        Metric & Purpose \\
    \midrule
        APO & Action Probability per Observation \\
        APD & Action Probability Distribution \\
    \midrule
        HAR & Hazard Action Ratio \\
        CAR & Constant-Pressure Action Ratio \\
    \midrule
        SPM & High Risk Scenarios per Million Flights \\
        FPM & Faults per Million Flights \\
    \bottomrule
    \end{tabular}}
    \caption{\textbf{Metrics for online evaluation of testing performance.} The metrics are categorized into three groups for a comprehensive evaluation of the proposed testing framework's capabilities, including the preference and quality of proposed framework, as well as the final results. The detail definition of metrics can be found in Appendix \ref{sec:metric_appendix}.}
    \label{tab:metric}
\end{table}

\paragraph{Action Probability per Observation (APO)}
The definition of APO is
\begin{equation*}
    \text{APO} = \frac{\#\{\text{action generated as injected, testing method is called}\}}{\#\{\text{testing method is called}\}} \times 100\%,
\end{equation*}
where $\#\{\cdot\}$ denotes the number of occurrences of the specified event.
APO aims to measure the percentage of times a testing method generates actions that are injected into the system, indicating how often the framework effectively targets the desired action space during testing.
However, high APO may result in redundant action injections, as not all injected actions contribute to uncovering valuable information. Only critical actions that can reveal faults or vulnerabilities are truly significant for effective testing.
Therefore, additional metrics about action quality and testing efficiency are necessary to evaluate the true effectiveness of the testing framework.

\paragraph{Action Probability Distribution (APD)} APD measures the proportion of different types of actions generated by the testing framework. It is represented as a vector indicating the percentage of each action type. A balanced APD ensures that the framework explores a diverse set of actions, while an unbalanced distribution may indicate bias toward specific types, potentially missing critical scenarios. Evaluating APD helps assess whether the testing method maintains comprehensive action coverage or if certain action types are underrepresented.

\paragraph{Hazard Action Ratio (HAR)} HAR is defined as
\begin{equation*}
    \text{HAR} = \frac{\#\{\text{actions result in return-to-go significantly raise comparing with summary}\}}{\#\{\text{injected actions}\}} \times 100\%,
\end{equation*}
where $\#\{\cdot\}$ denotes the number of occurrences of the specified event.
In practice, we consider an action to be hazardous if the difference between \textit{return-to-go} and the \textit{summary} is greater than 0.4.
This threshold indicates that the injected action has a substantial impact on the system, potentially leading to risky or unexpected outcomes.
A high HAR reflects the framework's ability to generate high-risk scenarios, which is crucial for identifying critical vulnerabilities during testing.

\paragraph{Constant-Pressure Action Ratio (CAR)} CAR is defined as
\begin{equation*}
    \text{CAR} = \frac{\#\{\text{actions result in high return-to-go when summary is also high}\}}{\#\{\text{injected actions}\}} \times 100\%,
\end{equation*}
where $\#\{\cdot\}$ denotes the number of occurrences of the specified event.
In practice, an action is categorized as constant-pressure if both the return-to-go and the summary exceed a threshold of 0.4.
This indicates that the action consistently maintains a high level of risk or pressure in an already high-risk scenario.
A high CAR shows that the testing framework is able to sustain pressure over a prolonged period, making it more effective at evaluating the resilience and stability of the system under stress.

\paragraph{High Risk Scenarios per Million Flights (SPM)}
SPM measures the frequency of high-risk scenarios detected by the testing framework for every million simulated flights.
A high SPM value indicates that the testing framework is effective in uncovering critical situations that pose potential threats to system safety.
It helps quantify the robustness of the testing methodology in identifying rare but impactful scenarios.

\paragraph{Faults per Million Flights (FPM)}
FPM represents the number of unique bugs identified for every million flights, where system may encounter severe failures.
It reflects the framework's capability to discover actual system faults during testing.
A higher FPM suggests that the testing strategy is not only triggering risky scenarios but also exposing underlying system vulnerabilities that need to be addressed before deployment.

\section{Architecture and Training Details}

\paragraph{Architectures of Policy Model}
The scenario-oriented testing framework for UTM systems consists of two main phases: training and inference (testing), as illustrated in Algorithms \ref{alg:training} and \ref{alg:inference}. Algorithm \ref{alg:training} details the training phase, where the Policy Model (PM) learns from an offline dataset of UTM scenarios. This phase involves iterating through epochs and batches, processing state-action-reward tuples, and updating the model parameters to minimize the prediction error for both actions and rewards. The training process incorporates context augmentation to enhance the model's ability to capture temporal dependencies. Algorithm \ref{alg:inference} outlines the inference (testing) phase, where the trained PM is used to generate and evaluate potentially vulnerable scenarios in the System-Under-Test (SUT). This phase operates in a loop, continuously generating candidate actions, filtering them through an Action Sampler (AS), injecting selected actions into the SUT, and evaluating the outcomes. The process accumulates detected vulnerabilities while dynamically updating the context based on observed states, actions, and rewards. Together, these algorithms form a comprehensive approach to identifying potential faults and vulnerabilities in UTM systems, leveraging both historical data and adaptive, context-aware scenario generation.

\begin{algorithm}
\caption{Training Phase of UTM Testing Framework}
\label{alg:training}
\begin{algorithmic}[1]
\Require{Offline dataset $D$, Model architecture $M$}
\Ensure{Trained Policy Model PM}

\State Initialize PM with architecture $M$
\State Initialize optimizer
\For{each epoch}
    \For{each batch $B$ in $D$}
        \State $s, a, r \gets$ GetBatchData($B$)
        \State $\tilde{s} \gets$ AugmentWithContext($s$)
        \State $\hat{a}, \hat{r} \gets$ PM.Forward($\tilde{s}$)
        \State $L \gets$ ComputeLoss($\hat{a}, a, \hat{r}, r$)
        \State BackpropagateAndUpdate(PM, $L$)
    \EndFor
\EndFor

\Return PM
\end{algorithmic}
\end{algorithm}

\begin{algorithm}
\caption{Inference (Testing) Phase of UTM Testing Framework}
\label{alg:inference}
\begin{algorithmic}[1]
    \Require{Trained Policy Model PM, System-Under-Test SUT, Action Sampler AS}
    \Ensure{Detected vulnerabilities $V$}
    \State Initialize vulnerability set $V \gets \emptyset$
    \State Initialize context set $C \gets \emptyset$
    \While{testing budget not exhausted}
        \State $s \gets$ GetCurrentState(SUT)
        \State $\tilde{s} \gets [C; s]$  \Comment{Augment state with context}
        \State $R_{predicted} \gets$ PM.PredictRTG($\tilde{s}$)
        \State $a_{candidates}, \gets$ PM.GenerateActions($\tilde{s}, R_{predicted}$)
        \State $a_{filtered} \gets$ AS.FilterActions($a_{candidates}$)
        \State $a \gets$ AS.SampleAction($a_{filtered}$)
        \State InjectAction(SUT, $a$)
        \State $R_{actual} \gets$ EvaluateAction(SUT, $a$)
        \If{IsVulnerability($r_{actual}$)}
            \State $V \gets V \cup \{(s, a, r_{actual})\}$
        \EndIf
        \State UpdateContext($C$, $s$, $a$, $r_{actual}$)
    \EndWhile
    \Return $V$
\end{algorithmic}
\end{algorithm}

\begin{table}[ht]
    \centering
    \begin{tabular}{ccc}
    \toprule
      &  PM-1.2B & PM-2B \\
    \midrule
     Layers & 64 & 64 \\
     Model Dimension & 1280 & 1600 \\
     Attention Heads & 20 & 25 \\
     Activation Functions & \multicolumn{2}{c}{GELU} \\
     Positional Embeddings& \multicolumn{2}{c}{Sinusoidal} \\
     \midrule
     Optimizer & \multicolumn{2}{c}{AdamW} \\
     Peak Learning Rate & $8\times 10^{-4}$ & $3\times 10^{-4}$ \\
     Learning Rate Schedule & \multicolumn{2}{c}{1000 steps warmup \& cosine decay} \\
     \midrule
     Batch Size & 512 & 256\\
     GPUs & \multicolumn{2}{c}{16} \\
    \bottomrule
    \end{tabular}
    \caption{\textbf{Overview of the key hyperparameters of policy model.} We display settings for 1.2B and 2B models.}
    \label{tab:model_param}
\end{table}

\paragraph{Action Space}
\label{Sec:App_action_space}
Considering feasibility in implementation, we defined the action space of PM with 2 types of actions: (1) \textbf{O}ne-time physical actions and (2) short-\textbf{D}uration digital actions. As shown in Table \ref{action_types_of_policy_model}, PM is also enabled to generate scenario configurations with different parameter settings.

\begin{table}[ht]
\begin{center}
\begin{tabular}{cccc}
\toprule
\multicolumn{1}{c}{\textbf{NAME}} &\multicolumn{1}{c}{\textbf{TYPE}} &\multicolumn{1}{c}{\textbf{ DESCRIPTION}} &\multicolumn{1}{c}{\bf PARAMETERS}
\\ \midrule
Wind &\textbf{O} &Winds with the exceeding magnitude &Speed, Direction\\
Obstacle &\textbf{O} &Obstacles appearing in UAVs' routes &Size, Location\\
Network Jitter &\textbf{D} &Temporary network disconnection &Time Duration\\
\bottomrule
\end{tabular}
\end{center}
\caption{\textbf{Action types of policy model.} We consider three types of action for each agent. The \textbf{O} stands for \textbf{O}ne-time physical actions and \textbf{D} stands for short-\textbf{D}uration digital actions.}
\label{action_types_of_policy_model}
\end{table}

\paragraph{Loss function}
We made use of model with decision transformer style which had out-standing in sparse reward tasks \citep{BhargavaWhenshouldwepreferDecisionTransformersOfflineReinforcementLearning2023}. In favor of regression of PM, a multi-objective loss function is introduced in training consisting of following aspects with configurable weights: \textit{return-to-go} to model observation and causality, \textit{action mask} to model world background knowledge and \textit{action} to model decision.

\section{Industry Level UAV Swarm Simulator}
The industry level UAV swarm simulator we applied is designed to create a digital twin of drone swarms for accurate analysis of both UTM system and UAVs' behaviors in real-world environments and interactions between natural environment and the whole system. Powered by a physics engine, the simulator closely replicates real-world physics. Additionally, the simulator incorporates hardware-in-the-loop by integrating actual UAV flight control systems, which adds to the accuracy. The simulator supports a variety of environmental configurations, including buildings, moving objects like balloons and birds, lighting conditions, and wind effects, etc. Backed by a dedicated support team, the system's reliability can be continuously improved.

\section{Environment Details}
\label{sec:envrionment_details}

\begin{figure}[ht]
\begin{center}
\begin{subfigure}[htbp]{0.4\textwidth}
    \begin{center}
    \includegraphics[height=4cm]{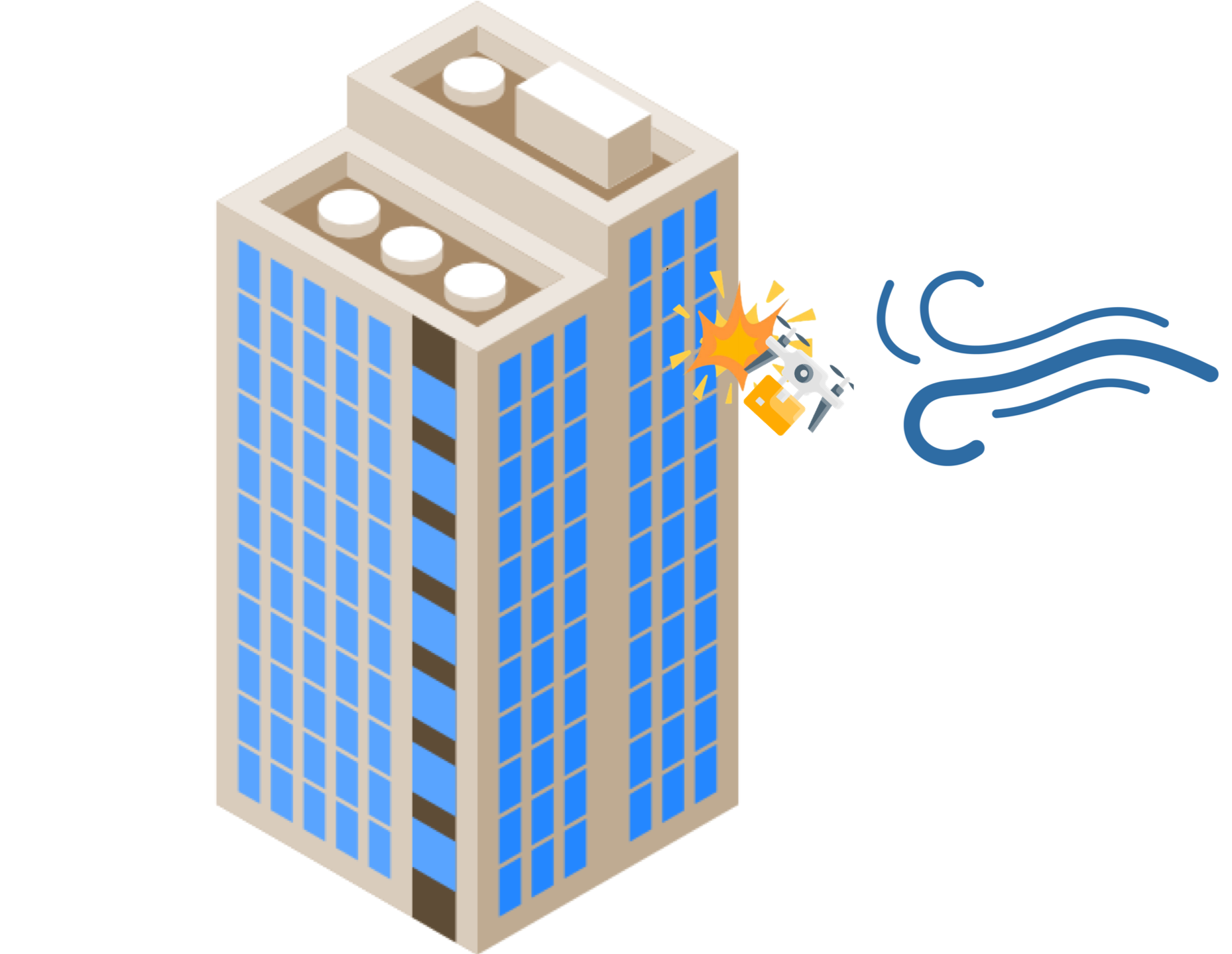}
    \end{center}
    \caption{Physical failures;}
    \label{fig:physic_failure}
\end{subfigure}
\begin{subfigure}[thb]{0.4\textwidth}
    \begin{center}
    \includegraphics[height=4cm]{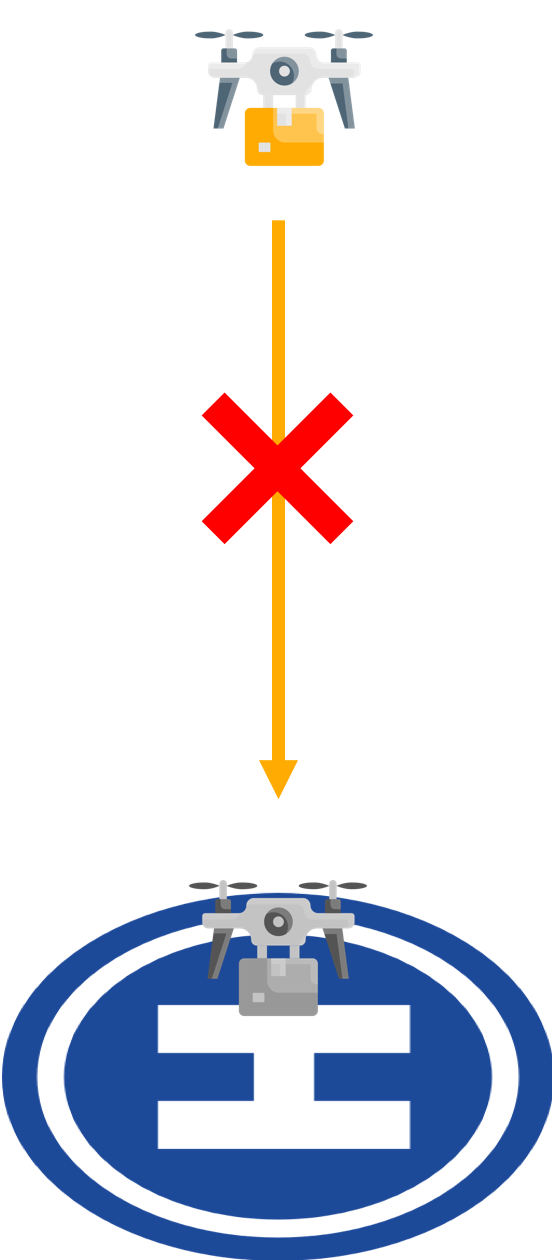}
    \end{center}
    \caption{Task failures;}
    \label{fig:task_failure}
\end{subfigure}
\end{center}
\caption{\textbf{Two main types of failures in UTM.} Physical failures: Failures that result from physical damage or malfunction in system components, such as structural damage, hardware breakdowns, or external impact. These failures typically require immediate attention as they compromise the safety and integrity of the UAV or surrounding environment. Task Failures: Failures related to mission objectives, such as incorrect task execution, navigation errors, etc. Task failures impact the operational success and can disrupt planned missions or lead to unexpected behavior.}
\label{fig:UAV_failures}
\end{figure}

\begin{table}[htbp]
    \centering
    \resizebox{\textwidth}{!}{
    \begin{tabular}{ccccccccc}
    \toprule
        Type & Index & Area & \# of Airport & \# of UAV & \# of Flight Line & \# of Alternate Airport & Fraction\\
    \midrule
        Offline Training & R1 & Rural  & 6 & 16 & 12 & 2 & 12.2\% \\
        Offline Training & R2 & Suburb & 12 & 24 & 24 & 7 & 18.3\% \\
        Offline Training & R3 & Urban & 6 & 36 & 18 & 6 & 27.5\% \\
        Offline Training & R4 & Suburb & 10 & 15 & 10 & 2 & 11.5\% \\
        Offline Training & R5 & Suburb & 10 & 15 & 10 & 2 & 9.2\% \\
        Offline Training & R6 & Urban & 8 & 16 & 16 & 2 & 12.2\% \\
        Offline Training & R7 & Urban & 4 & 12 & 8 & 3 & 9.1\% \\
    \midrule
        Online Testing & TR1 & Rural & 9 & 29 & 16 & 2 & N/A \\
        Online Testing & TR2 & Urban & 6 & 16 & 16 & 6 & N/A \\
    \bottomrule
    \end{tabular}}
    \caption{\textbf{Overview of training and testing regions used in the scenario-based testing framework.} Each region is categorized by type (rural, suburban, or urban) and is characterized by attributes such as the number of airports, UAVs, flight lines, and alternate airports. For training dataset, the fraction of each region is provided to reflect the distribution of different operational environments. Each region is specifically designed to provide a representative mix of operational challenges: regions R1 and R4 emphasize low-density rural and suburban operations, respectively, whereas regions R3 and R6 represent high-density urban areas with increased air traffic complexity. This distribution ensures the model learns to generalize across different environment types while prioritizing scenarios with a higher likelihood of critical interactions. Testing regions are designed to evaluate model performance on both trained dataset and unseen scenarios, ensuring robustness and generalizability.}
    \label{tab:region}
\end{table}

\end{document}